\documentclass{bmvc2k}

\newcommand{\inst}[1]{\textsuperscript{#1}}

\title{In-Model Merging for Enhancing the Robustness of Medical Imaging Classification Models}

\addauthor{Hu Wang\inst{1}, Ibrahim Almakky\inst{1}, \\
Congbo Ma\inst{2}, Numan Saeed\inst{1}, \\
Mohammad Yaqub}{}{1}

\addinstitution{
Mohamed bin Zayed University \\
of Artificial Intelligence, UAE
}
\addinstitution{
New York University Abu Dhabi, UAE
}

\runninghead{Hu Wang et al.}{In-Model Merging}

\usepackage[ruled,vlined]{algorithm2e}
\usepackage{graphicx}
\usepackage{multirow}
\usepackage{xcolor}
\usepackage{pifont}
\usepackage{amsmath}

\usepackage{algorithmic}
\SetKwInput{KwIn}{Input}
\SetKwInput{KwOut}{Output}

\begin{document}

\maketitle

\begin{abstract}
Model merging is an effective strategy to merge multiple models for enhancing model performances, and more efficient than ensemble learning as it will not introduce extra computation into inference. However, limited research explores if the merging process can occur within one model and enhance the model's robustness, which is particularly critical in the medical image domain. In the paper, we are the first to propose in-model merging (InMerge), a novel approach that enhances the model's robustness by selectively merging similar convolutional kernels in the deep layers of a single convolutional neural network (CNN) during the training process for classification. We also analytically reveal important characteristics that affect how in-model merging should be performed, serving as an insightful reference for the community. We demonstrate the feasibility and effectiveness of this technique for different CNN architectures on 4 prevalent datasets. The proposed InMerge-trained model surpasses the typically-trained model by a substantial margin. The code will be made public.
\end{abstract}

\section{Introduction}

Deep learning has become the cornerstone of modern machine learning. The performance of deep convolutional models is heavily dependent on the design and optimization of convolutional neural networks, which hierarchically extract features from input data. Ensemble learning is an effective manner for enhancing the model performances of medical imaging analysis (MIA), where tasks such as disease diagnosis demand high levels of accuracy. However, the inference computation will increase linearly.

Model merging, on the other hand, aims to improve performance without extra inference cost by combining weights from multiple trained models. Methods like Model Soups~\cite{wortsman2022model,wang2024rethinking} and Ties-merging~\cite{yadav2023ties} demonstrate improved generalization by averaging or resolving weight conflicts. Other approaches view merging as interpolating over loss basins~\cite{ainsworth2022git,entezari2021role,jordan2022repair,stoica2023zipit}. However, these are inter-model methods that require multiple trained models.
An important research question remains and limited research has explored it: can we conduct merging operations within a single model to enhance model robustness for free and without introducing extra models?

To answer this question and bridge the gap, in this work, we propose In-model Merging, a novel technique that selectively merges similar convolutional kernels within a single model. \textit{To the best of our knowledge, we are the first to consider merging techniques for a single model.} By strategically merging similar kernels, the pattern redundancy between kernels can be diminished, a regularization effect will thus contribute to enhance the model's robustness. We aim to enhance the model's robustness and improve its capability to represent features more effectively. The similarity of kernels is determined by cosine similarity.
We analytically demonstrate that shallow layers are crucial for low-level feature extraction and inappropriate for in-model merging. Also, we identify suitable similarity thresholds and merge weights/probabilities with extensive analysis experiments. By dynamically merging convolutional kernels, our method improves feature representation and model performance. 
Our work is generic, but it has important applications in MIA, which is our focus in the paper and where model robustness is essential. Our contributions are summarized as follows:

\begin{itemize}
\item We propose a novel In-model Merging (InMerge) technique for enhancing the robustness of medical imaging classification models. To the best of our knowledge, we are the first to consider merging techniques from single model perspective.

\item We enhance the model performance with minimal cost by proposing a In-model Merging finetuning process, where similar convolutional kernels within one single model are dynamically merged based on their pair-wise cosine similarities during training. It can serve as a plug-and-play module for different CNN architectures.

\item We demonstrate the effectiveness of the proposed In-model Merging in the challenging context of MIA. Through extensive analysis, we find that restricting the merging process to deeper layers while keeping low-level feature extraction intact leads to better model performance, as low-level features are prone to be influenced by merging kernels.
This paper also reveals other important characteristics that affect how In-model Merging should be performed to get useful insights.
\end{itemize}

\section{Related Work}

Model merging has recently gained attention for its ability to improve model performance without introducing extra computation in the inference. By averaging weights from different models, Model Soups~\cite{wortsman2022model,wang2024rethinking} can lead to improved generalization and robustness with simply merging models. Ties-merging~\cite{yadav2023ties} takes one step ahead by taking weight conflicts into account while merging. There are multiple models perform merging from a different perspective by considering the merging process of finding a good way to interpolate on the loss basins~\cite{ainsworth2022git,entezari2021role,jordan2022repair,stoica2023zipit}. Existing works~\cite{wortsman2022model,ainsworth2022git,yadav2023ties,chen2023fedsoup,almakky2024medmerge,stoica2023zipit} have demonstrated the potential of merging techniques to optimize neural network performance across multiple models. However, these techniques are inter-model, requiring multiple models trained for merging.

\section{Methodology}

To improve the robustness of feature representation in CNNs, we propose a novel In-model Merging strategy. The model selectively merges similar convolutional kernels in the deeper layers of the model, while preserving the integrity of shallow layers. By merging kernels and reducing the kernel redundancies, the approach enhances model performance. It is worth noting that the proposed in-model merging method works as a finetuning approach. After a model is pretrained, In-model Merging requires a few epochs of finetuning, which ensures it will not introduce much computation overhead during the training process. In addition, no extra computation is added at inference time, as there will not be any merging operations at inference.

\subsection{Kernel Similarity Computation}

\begin{figure}
\centering
\includegraphics[width=0.8\textwidth]{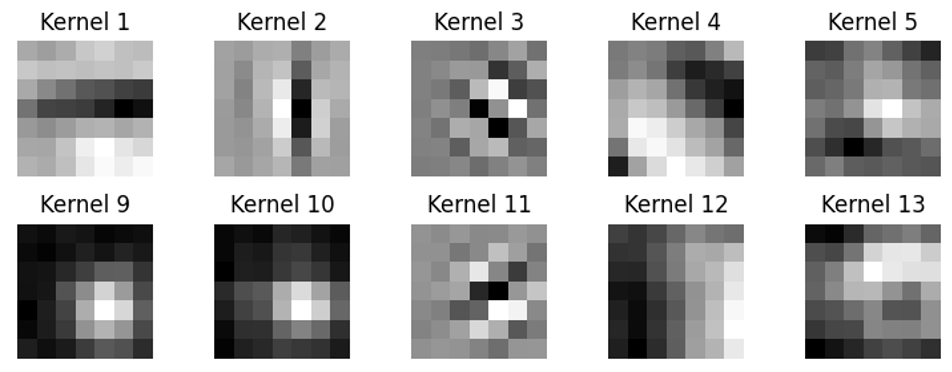}
\caption{An example showing the similarity differences between patterns stored in kernels in a well-trained ResNet34 layer. The more similar pair of kernels would incur larger similarity. Kernel 9 and Kernel 10 are similar, the similarity $\text{sim}(\mathbf{k}_9, \mathbf{k}_{10})=0.9372$; Kernel 3 and Kernel 11 nearly have orthogonal textures, so $\text{sim}(\mathbf{k}_3, \mathbf{k}_{11})=-0.0234$; Kernel 4 and Kernel 12 have very different textures and colors, $\text{sim}(\mathbf{k}_4, \mathbf{k}_{12})=-0.5049$; Kernel 5 and Kernel 13 are somewhat similar, $\text{sim}(\mathbf{k}_5, \mathbf{k}_{13})=0.4562.$} \label{fig:conv_kernels}
\end{figure}

The similarity between two convolutional kernels is measured using cosine similarity, which quantifies the angular difference in their vectorized representations. Given two kernels $\mathbf{K}_i$ and $\mathbf{K}_j$ from the same convolutional layer, we reshape them into one-dimensional vectors:
\begin{equation}
\mathbf{k}_i = \text{vec}(\mathbf{K}_i), \quad \mathbf{k}_j = \text{vec}(\mathbf{K}_j),
\end{equation}
where $\text{vec}(\cdot)$ denotes the vectorization operation that flattens the kernel into a column vector. The cosine similarity is then computed as:
\begin{equation}
\text{sim}(\mathbf{k}_i, \mathbf{k}_j) = \frac{\mathbf{k}_i^T \mathbf{k}_j}{{\|\mathbf{k}_i\| \|\mathbf{k}_j\|}},
\end{equation}
where $||\cdot||$ represents the $\ell_2$-norm. This metric ensures that similarity is independent of the absolute magnitudes of the kernels, focusing solely on their directional alignment.

\subsection{In-Model Kernel Merging Strategy}

To preserve the integrity of shallow layer features, the merging process is constrained to be only performed on deeper layers. In those convolutional layers, beyond the first $L_s$ shallow layers, similar kernels are merged dynamically during training. Let $\mathbf{W} = {\mathbf{K}_1, \mathbf{K}_2, \dots, \mathbf{K}_n}$ denote the set of kernels in a convolutional layer. For each kernel $\mathbf{K}_i$, another kernel $\mathbf{K}_j$ is randomly selected such that $j \neq i$. If the similarity criterion is met, the two kernels are merged using a weighted interpolation with a certain probability $p$:
\begin{equation}
\mathbf{K}_i \leftarrow \alpha \mathbf{K}_i + (1-\alpha) \mathbf{K}_j, \;\;s.t. \; \text{sim}(\mathbf{k}_i, \mathbf{k}_j) > \tau,
\end{equation}
where $\tau \in [-1,1]$ is the similarity threshold. 
This merging operation helps mitigate redundancy in feature extraction while preserving the network expressiveness.

\subsection{Layer Selection and Probabilistic Merging}

To maintain the diversity of low-level feature representations, the first $L_s$ shallow layers are excluded from kernel merging. Additionally, merging is performed in a probabilistic manner to avoid excessive regularization and unintended degradation of feature maps. Specifically, each kernel undergoes merging with probability $p$, ensuring stochasticity in the process:

\begin{equation}
P(\mathbf{K}_i \leftarrow \alpha \mathbf{K}_i + (1-\alpha) \mathbf{K}_j) = p.
\end{equation}

This probabilistic merging mechanism prevents the network from over-collapsing to a limited set of feature extractors, giving the effect of regularization, thereby retaining sufficient expressiveness while enhancing generalization.

\subsection{Algorithm and Rationale}
The kernel merging process is repeated at each training iteration. The In-model Merging algorithm is shown as follows:

\begin{algorithm}[H]
\caption{In-model Merging for Enhancing the Robustness of Medical Imaging Classification Models}
\label{alg:kernel_merging}
\KwIn{Loaded pretrained model $\mathcal{M}$ with convolutional layers $\{\mathcal{L}_1, \mathcal{L}_2, \dots, \mathcal{L}_N\}$, shallow layer count $L_s$, merging probability $p$, similarity threshold $\tau$.}
\KwOut{Updated model $\mathcal{M}$ with in-model merging.}

\For{each training iteration}{
    \For{each convolutional layer $\mathcal{L}_i$ in $\mathcal{M}$}{
        \If{$i < L_s$}{
            \textbf{continue};
        }

        Retrieve kernel weights $\mathbf{W} = \{\mathbf{K}_1, \mathbf{K}_2, \dots, \mathbf{K}_n\}$\;
        Initialize new weights $\mathbf{W}^\prime \gets \mathbf{W}$\;

        \For{each kernel $\mathbf{K}_i \in \mathbf{W}$}{
            With probability $p$, randomly select another kernel $\mathbf{K}_j$ such that $j \neq i$\;

            Compute cosine similarity:
            \[
            \text{sim}(\mathbf{k}_i, \mathbf{k}_j) = \frac{\mathbf{k}_i \cdot \mathbf{k}_j}{\|\mathbf{k}_i\| \|\mathbf{k}_j\|}
            \]

            \If{$\text{sim}(\mathbf{k}_i, \mathbf{k}_j) > \tau$}{
                Merge kernels:
                \[
                \mathbf{K}_i \leftarrow \alpha \mathbf{K}_i + (1-\alpha) \mathbf{K}_j
                \]
                Update $\mathbf{W}^\prime[i] \gets \mathbf{K}_i$\;
            }
        }

        Update layer weights via back-propagation: $\mathcal{L}_i.\mathbf{W} \gets \mathbf{W}^\prime$\;
    }
}
\end{algorithm}

As an example shown in Fig.~\ref{fig:conv_kernels}, the similar kernel pair has a larger similarity, while the dis-similar kernel pair has a smaller similarity. The rationale of why it can work is that there are patterns stored in learned weights. Convolutional kernels in different layers often capture overlapping or similar patterns, which introduce redundancy in learned features. When merging the similar patterns, it reduces redundant components within one single model. The process therefore has the same effect of regularization, so that the redundancy can be suppressed and model robustness can be enhanced. This selective and iterative merging approach ensures the model maintains a balance between feature diversity and redundancy reduction.


\section{Experiments}

\subsection{Datasets and Implementation Details}

\subsubsection{ChestXRay14 Tasks}

ChestXRay-14 is a chest X-ray imaging dataset for multi-label classification. It contains 112,120 frontal chest X-ray images from 30,805 unique patients. The dataset is labeled with 14 different lung disease categories, including conditions of Atelectasis, Nodule, Pneumonia, Edema and others. These labels are based on radiologist annotations, and each image can have one or more labels associated with it, making it a multi-label classification task.
For the dataset, we adopt the official split for training and evaluation. For model training, we apply a series of standard data augmentations and transformations. These include random horizontal flipping, resizing the images to 224 by 224 pixels. Additionally, the images are converted into tensors and normalized to ensure consistent input scaling. 
The model is trained with a sigmoid head on 14 outputs and using 14 Binary Cross-Entropy Losses for the multi-label classification task. The optimizer chosen for the training process is Stochastic Gradient Descent (SGD) with a staged learning rate strategy and an initial learning rate of 0.01, momentum set to 0.9, and weight decay of 1e-4 for regularization. The batch size is set to 324. The default backbone network is based on ImageNet pretrained VGG19 model. After the backbone model is trained (by default 30 epochs), an extra 10 epochs for in-model merging finetuning is applied.

\subsubsection{MedMNIST Tasks}

MedMNIST data is more diverse to the chestxray data. PathMNIST, DermaMNIST, and OCTMNIST are in the collection of MedMNIST and commonly used in MIA. 
\textbf{PathMNIST} is a dataset of histopathological images with 9 classes, particularly focused on classifying tissue samples, such as breast cancer tissue, into categories like malignant or benign. It is designed for evaluating models that can automatically analyze and categorize tissue samples from whole-slide scans. The sample number is 107,180 (89,996 for training, 10,004 for validation and 7,180 for testing). 
\textbf{DermaMNIST}, on the other hand, contains dermoscopic images of skin lesions with 7 classes and is typically used for the classification of various skin diseases, such as melanoma and non-melanoma types. This dataset is valuable for models aimed at automating the diagnosis of skin conditions based on visual cues from lesions. The sample number is 10,015 (7,007 for training, 1,003 for validation and 2,005 for testing). 
\textbf{BloodMNIST} BloodMNIST is a medical image classification dataset consisting of 17,092 blood cell images across 8 classes. The dataset is split into 11,959 training images, 1,712 validation images, and 3,421 test images. It is derived from a larger dataset of peripheral blood smear images. 
The implementation fine-tunes a pre-trained VGG19 model on the MedMNIST dataset. Data preprocessing includes resizing images to 64 by 64 and using a batch size of 1024. Training (20 epochs backbone training and 5 epochs InMerge training) is performed using SGD with a multi-step learning rate schedule. The model's performance is evaluated on a validation set each epoch, with the best-performing model saved. In the experiments, each model is trained and tested 5 times to fetch its mean and standard deviation.

For all these datasets, we generally set merging weight $\alpha$ to 0.8; skipped layer $L_s$ to 10 for VGG19 as default; merging probability $p$ to 0.3 and similarity threshold $\tau$ to 0.3. In-model Merging is abbreviated as InMerge in the following figures/tables/descriptions. To maintain a fair comparison, the baseline models on all datasets are trained with the same number of epochs as the InMerge model, and we focus our analysis on CNN models.

\subsection{Results \& Discussion}

\begin{table}[h!]
\setlength\tabcolsep{1pt}
    \centering
    \resizebox{1.0\textwidth}{!}{
    \begin{tabular}{l|cccccccccccccc|c}
        \hline \hline
        Models & Atel & Card & Effu & Infi & Mass & Nodu & Pneu1 & Pneu2 & Cons & Edem & Emph & Fibr & P\_T & Hern & Ave \\
        \hline
        U-DCNN~\cite{wang2017chestx} & 0.700 & 0.810 & 0.759 & 0.661 & 0.693 & 0.669 & 0.658 & 0.799 & 0.703 & 0.805 & 0.833 & 0.786 & 0.684 & 0.872 & 0.745 \\
        LSTM~\cite{yao1710learning} & 0.733 & 0.856 & 0.806 & 0.673 & 0.718 & 0.777 & 0.684 & 0.805 & 0.711 & 0.806 & 0.842 & 0.743 & 0.724 & 0.775 & 0.761 \\
        AGCL~\cite{tang2018attention} & 0.756 & 0.887 & 0.819 & 0.689 & 0.814 & 0.755 & 0.729 & 0.850 & 0.728 & 0.848 & 0.906 & 0.818 & 0.765 & 0.875 & 0.803 \\
        CheXNet~\cite{rajpurkar2017chexnet} & 0.769 & 0.885 & 0.825 & 0.694 & 0.824 & 0.759 & 0.715 & 0.852 & 0.745 & 0.842 & 0.906 & 0.821 & 0.766 & 0.901 & 0.807 \\
        DNet~\cite{guendel2019learning} & 0.767 & 0.883 & 0.828 & 0.709 & 0.821 & 0.758 & 0.731 & 0.846 & 0.745 & 0.835 & 0.895 & 0.818 & 0.761 & 0.896 & 0.807 \\
        CRAL~\cite{guan2020multi} & 0.781 & 0.880 & 0.829 & 0.702 & 0.834 & 0.773 & 0.729 & 0.857 & 0.754 & 0.850 & 0.908 & 0.830 & 0.778 & 0.917 & 0.816 \\
        CAN~\cite{ma2019multi} & 0.777 & 0.894 & 0.829 & 0.696 & 0.838 & 0.771 & 0.722 & 0.862 & 0.750 & 0.846 & 0.908 & 0.827 & 0.779 & \textbf{0.934} & 0.817 \\
        DualCXN~\cite{chen2019dualchexnet} & 0.784 & 0.888 & 0.831 & 0.705 & 0.838 & 0.796 & 0.727 & 0.876 & 0.746 & 0.852 & 0.942 & \textbf{0.837} & 0.796 & 0.912 & 0.823 \\
        LLAGnet~\cite{chen2019lesion} & 0.783 & 0.885 & 0.834 & 0.703 & 0.841 & 0.790 & 0.729 & 0.877 & 0.754 & 0.851 & 0.939 & 0.832 & \textbf{0.798} & 0.916 & 0.824 \\
        CheXGCN~\cite{chen2020label} & 0.786 & 0.893 & 0.832 & 0.699 & 0.840 & \textbf{0.800} & 0.739 & 0.876 & 0.751 & 0.850 & \textbf{0.944} & 0.834 & 0.795 & 0.929 & 0.826 \\ \hline
        InMerge (ours) & \textbf{0.805} & \textbf{0.895} & \textbf{0.881} & \textbf{0.710} & \textbf{0.842} & 0.750 & \textbf{0.770} & \textbf{0.878} & \textbf{0.801} & \textbf{0.888} & 0.915 & 0.818 & 0.773 & 0.903 & \textbf{0.830} \\
        \hline \hline
    \end{tabular}}
    \caption{Performance Comparison of Different Models in AUROC. The 14 different lung disease categories are Cardiomegaly, Effusion, Infiltration, Mass, Nodule, Pneumonia, Pneumothorax, Consolidation, Edema, Emphysema, Fibrosis, Pleural Thickening, Hernia (the abbreviations in the table keep this order). The compared models are U-DCNN~\cite{wang2017chestx}, LSTM~\cite{yao1710learning}, AGCL~\cite{tang2018attention}, CheXNet~\cite{rajpurkar2017chexnet}, DNet~\cite{guendel2019learning}, CRAL~\cite{guan2020multi}, CAN~\cite{ma2019multi}, DualCheXNet (Abbr. DualCXN)~\cite{chen2019dualchexnet}, LLAGnet (Abbr. LLAGnet)~\cite{chen2019lesion} and CheXGCN~\cite{chen2020label}.
    ``DualCXN'' in the table denotes DualCheXNet.
    The best-performed model for each column is bolded.}
    \label{tab:chex_results}
\end{table}

\noindent\textbf{Chest X-Ray Tasks} As shown in Tab.~\ref{tab:chex_results}, the proposed InMerge model consistently outperforms existing techniques across multiple disease categories, achieving the highest average performance. Notably, it ranks first in 10 out of 15 ranks (including average), demonstrating its broad applicability and robustness. Particularly, InMerge exhibits large improvements in conditions such as Atelectasis (0.805), Effusion (0.881), and Consolidation (0.801) in AUC, suggesting that its ability to refine feature representation through in-model merging is highly beneficial for identifying complex patterns of these diseases. These results indicate that selectively merging similar convolutional kernels in deeper layers enhances representation learning with minimal computational burden, making the model more accurate.

Compared to prior methods, InMerge provides a noticeable improvement over current best architectures, e.g., CheXGCN, which shows strong performance in most cases but falls short in several other conditions. While CheXGCN benefits from graph-based relational modeling, InMerge consistently maintains high accuracy across diverse disease categories. Additionally, models such as CAN, DualCheXNet and LLAGnet, which integrate multi-branch or attention-based mechanisms, demonstrate competitive performance but do not consistently surpass InMerge, indicating that incorporating more parameters in the training may not always be the most effective strategy. Furthermore, when examining disease categories such as Pneumonia and Edema, InMerge also exhibits strong results. The strength of InMerge lies in its ability to dynamically refine internal representations by merging redundant yet functionally similar convolutional kernels, leading to improved generalization without the need for explicit architectural modifications or additional supervision.

Its superior performance in these categories suggests that merging similar feature detectors in deep layers enhances model robustness, reducing the risk of overfitting to spurious correlations often present in medical imaging datasets. Another critical observation is its competitive performance in Mass and  Fibrosis, where models such as CAN and DualCheXNet previously dominated. While methods like CAN leverage cross-model attention learning to enhance feature representations, InMerge achieves comparable results without requiring additional training objectives. These findings empirically show the idea that InMerge introduces an effective approach to optimizing convolutional networks for medical image classification. By reducing internal redundancy in learned features, InMerge enhances representation robustness without requiring additional data or architectural modifications. This makes it a promising candidate for improving deep learning models in medical imaging, where robustness and generalization are critical for real-world deployment.

\begin{table}[h!]
    \centering
    \resizebox{0.9\textwidth}{!}{
    \begin{tabular}{l|c c c|c c}
        \hline \hline
        \textbf{Dataset} & \textbf{Cls\#} & \textbf{Samp\#} & \textbf{Train/Val/Test} & \textbf{Baseline} & \textbf{InMerge (ours)} \\
        \hline
        PathMNIST & 9 & 107,180 & 89,996/10,004/7,180 & 0.898±0.010 & \textbf{0.924±0.002} \\
        DermaMNIST & 7 & 10,015 & 7,007/1,003/2,005 & 0.745±0.009 & \textbf{0.760±0.005} \\
        BloodMNIST & 8 & 17,092 & 11,959/1,712/3,421 & 0.941±0.005 & \textbf{0.948±0.004} \\
        \hline \hline
    \end{tabular}}
    \caption{Performance comparison of Baseline and InMerge across different MedMNIST datasets in accuracy.}
    \label{tab:medmnist_results}
\end{table}

\noindent\textbf{MedMNIST Tasks} In Tab.~\ref{tab:medmnist_results}, we present a comparative classification accuracy evaluation of our proposed In-Model Merging (InMerge) method against the baseline (VGG19 model \textit{without In-Model Merging plugged in}, trained with the same total number of epochs with the InMerge model) across multiple MedMNIST datasets, covering diverse medical imaging tasks. The results consistently demonstrate that InMerge outperforms the baseline across all datasets. These improvements are particularly noteworthy given that InMerge introduces minimal architectural modifications and computational costs. In terms of standard deviation, the results of In-model Merging across the three datasets consistently exhibit lower values compared to its baseline model, demonstrating greater stability across different training runs.

The best improvement is observed in PathMNIST, where InMerge achieves an average accuracy of 92.4\%, surpassing the baseline's 89.8\% by a large margin (2.9\% improvement). It suggests that our method effectively enhances feature representation in complex tissue patterns via regularization in convolutional kernels, leading to more robust feature extraction. On BloodMNIST, involving microscopic blood cell images, InMerge also yields improvement, despite not as noticable as on other two datasets. The enhancement on DermaMNIST (76.0\% vs. 74.5\%) is meaningful in dermatology applications where subtle textural differences are critical for classification. This suggests that even in datasets with limited training samples, InMerge merges redundant convolutional kernels to provide a regularization effect, reducing overfitting while preserving critical feature diversity. These results suggest the robustness of In-model Merging for medical image classification tasks. By selectively merging similar convolutional kernels in deep layers, InMerge enhances feature efficiency and model robustness without requiring additional resources. These findings highlight the potential of in-model merging as a practical and efficient strategy for improving medical image processing.

\subsection{Ablation Study and Analysis}

The experiments in this subsection are conducted on the chest X-ray data.

\subsubsection{Ablation: Merging with Different Weights $\alpha$ and Probabilities $p$}

\begin{figure}[h!]
\centering
\includegraphics[width=0.45\textwidth]{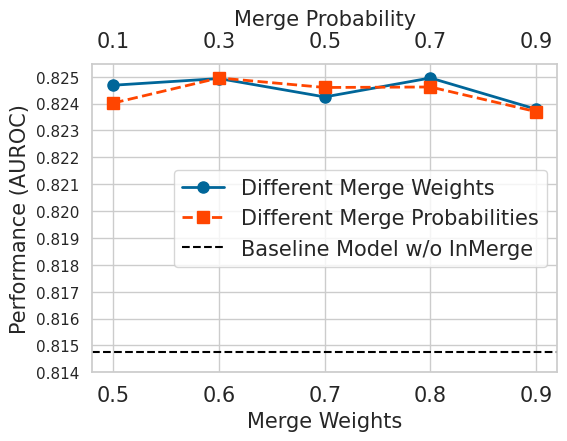}
\caption{Merging with different merge weights and merge probabilities}
\label{fig:merge}
\end{figure}

From Fig.~\ref{fig:merge}, we can observe from the perspective of either different merge weights $\alpha$ or probabilities $p$, the line plots show a stable trend and highly above the baseline model (VGG19 trained with the same number of epochs, but without InMerge operations), also serving as the ablation study of \textit{w and w/o InMerge}.

\subsubsection{Merging Layer $L_s$ Sensitivity}

\begin{figure}[h!]
\centering
\includegraphics[width=0.8\textwidth]{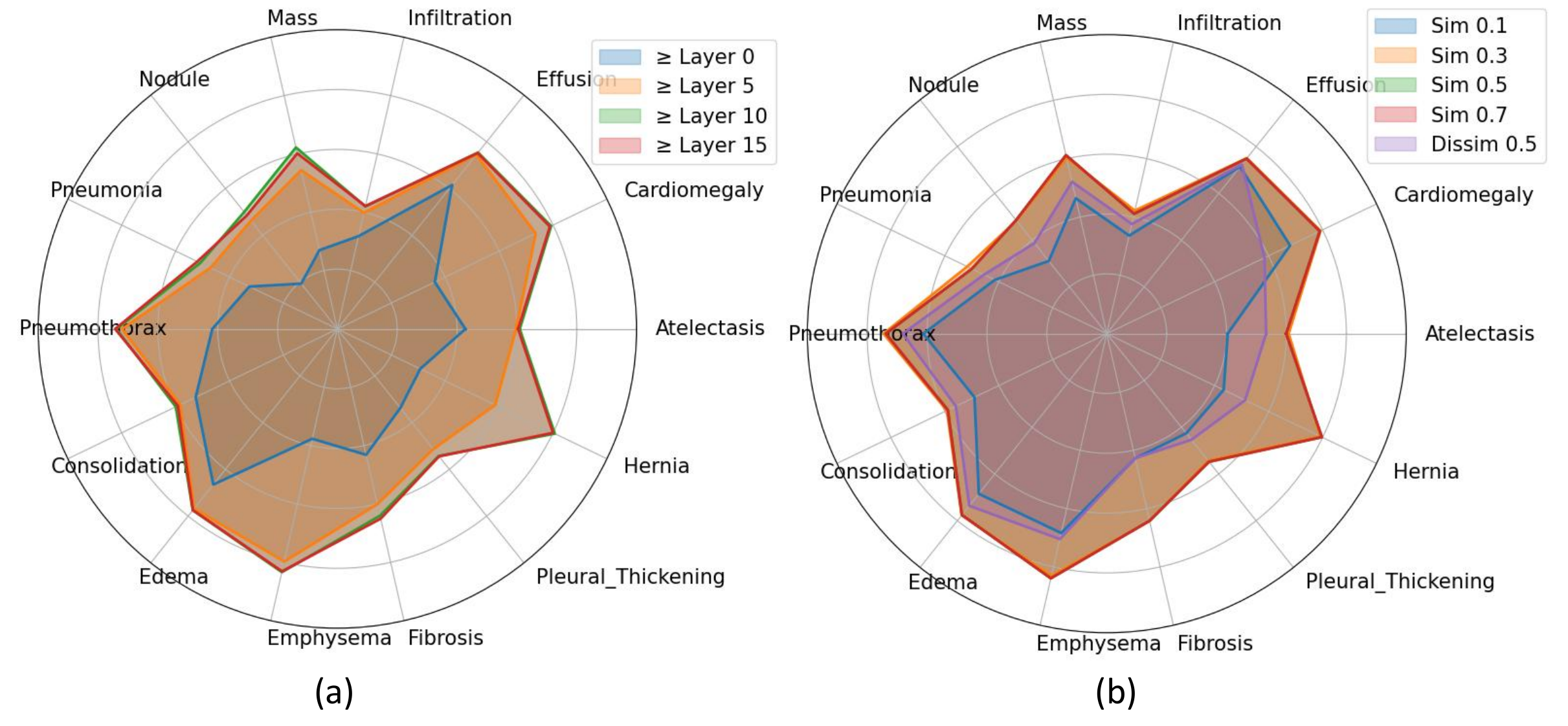}
\caption{(a) The sensitivity of different merged layers; (b) Merging with different similarities.}
\label{fig:layer_sim}
\end{figure}


Fig.~\ref{fig:layer_sim} (a) shows performance variations across merging different layers ($\geq$ Layer 10 means any layer below 10th layer is not considered for merging). A clear trend shows where only including deeper layers (Layer 10 and Layer 15) outperform including from shallower ones (Layer 0 and Layer 5) in most categories. This highlights the importance of deep feature representations for disease identification. Notably, conditions such as Emphysema, Cardio., and Pneumo. show substantial improvements when merging deeper layers, suggesting the possibility that these conditions rely more on low-level abstract features, so they are sensitive to layer change. In contrast, some categories, such as Infiltration and Consolidation, exhibit relatively stable or even slightly better performance at shallower layers, possibly due to their reliance on low-level texture features. 
For the setup of layer 0 and 5, the model performances are very different from layer 10, but models with $\geq$ layer 10 and 15 performance are similar, showing the general stability when layers go deeper. These findings empirically prove the idea that deeper feature merging is crucial for model merging strategies, while also indicating that the optimal layer depth for different diseases may vary, which is valuable for task-specific model merging setup.

\subsubsection{Merging with Different Similarities $\tau$}


Fig.~\ref{fig:layer_sim} (b) compares model performance under varying similarity thresholds for in-model kernel merging. A clear trend emerges where higher similarity thresholds (e.g., $\geq$ Sim 0.3) generally lead to stronger performance than the thresholds 0.1 or Dissimilar condition (we show dissimilar threshold 0.5 in the figure as well), where more dissimilar kernels are merged, indicating that merging dissimilar kernels disrupts critical feature representations. Interestingly, for Effusion and Edema, the performance difference between different similarity settings is relatively small, suggesting that these conditions may be less sensitive to in-model merging. These results highlight the importance of selective merging strategies, where preserving high-similarity features is beneficial while merging dissimilar kernels can degrade performance.

\subsubsection{In-model Merging on Different CNN Architectures}

We also examine the generality of In-model Merging on different models. For VGG16 model ($L_s=$10), the average AUC of In-model Merging surpasses the baseline model (0.817 VS. 0.807). Similar results are shown on ResNet34 ($L_s=$30), the model performance gained from 0.792 to 0.796.

\section{Conclusion}

We propose In-model Merging, where the merging process occurs within one model and enhances the model robustness with minimal training computations and no additional computations in inference. By selectively merging similar kernels in deeper layers, the method enhances feature representation while preserving low-level features for robustness. Our experiments confirm the effectiveness of in-model merging in improving model performance and robustness. Results across multiple medical datasets show improved generalization. This suggests that in-model merging is a practical enhancement for neural networks with potential for further exploration. Future research may focus on exploring InMerge in different network architectures, such as Transformers~\cite{vaswani2017attention} or Mamba~\cite{gu2023mamba}.

\bibliography{egbib}

\begin{thebibliography}{21}
\providecommand{\natexlab}[1]{#1}
\providecommand{\url}[1]{\texttt{#1}}
\expandafter\ifx\csname urlstyle\endcsname\relax
  \providecommand{\doi}[1]{doi: #1}\else
  \providecommand{\doi}{doi: \begingroup \urlstyle{rm}\Url}\fi

\bibitem[Ainsworth et~al.(2022)Ainsworth, Hayase, and Srinivasa]{ainsworth2022git}
Samuel~K Ainsworth, Jonathan Hayase, and Siddhartha Srinivasa.
\newblock Git re-basin: Merging models modulo permutation symmetries.
\newblock \emph{arXiv preprint arXiv:2209.04836}, 2022.

\bibitem[Almakky et~al.(2024)Almakky, Sanjeev, Hashmi, Qazi, and Yaqub]{almakky2024medmerge}
Ibrahim Almakky, Santosh Sanjeev, Anees Ur~Rehman Hashmi, Mohammad~Areeb Qazi, and Mohammad Yaqub.
\newblock Medmerge: Merging models for effective transfer learning to medical imaging tasks.
\newblock \emph{arXiv preprint arXiv:2403.11646}, 2024.

\bibitem[Chen et~al.(2019{\natexlab{a}})Chen, Li, Guo, and Lu]{chen2019dualchexnet}
Bingzhi Chen, Jinxing Li, Xiaobao Guo, and Guangming Lu.
\newblock Dualchexnet: dual asymmetric feature learning for thoracic disease classification in chest x-rays.
\newblock \emph{Biomedical Signal Processing and Control}, 53:\penalty0 101554, 2019{\natexlab{a}}.

\bibitem[Chen et~al.(2019{\natexlab{b}})Chen, Li, Lu, and Zhang]{chen2019lesion}
Bingzhi Chen, Jinxing Li, Guangming Lu, and David Zhang.
\newblock Lesion location attention guided network for multi-label thoracic disease classification in chest x-rays.
\newblock \emph{IEEE journal of biomedical and health informatics}, 24\penalty0 (7):\penalty0 2016--2027, 2019{\natexlab{b}}.

\bibitem[Chen et~al.(2020)Chen, Li, Lu, Yu, and Zhang]{chen2020label}
Bingzhi Chen, Jinxing Li, Guangming Lu, Hongbing Yu, and David Zhang.
\newblock Label co-occurrence learning with graph convolutional networks for multi-label chest x-ray image classification.
\newblock \emph{IEEE journal of biomedical and health informatics}, 24\penalty0 (8):\penalty0 2292--2302, 2020.

\bibitem[Chen et~al.(2023)Chen, Jiang, Dou, Wang, and Li]{chen2023fedsoup}
Minghui Chen, Meirui Jiang, Qi~Dou, Zehua Wang, and Xiaoxiao Li.
\newblock Fedsoup: improving generalization and personalization in federated learning via selective model interpolation.
\newblock In \emph{International Conference on Medical Image Computing and Computer-Assisted Intervention}, pages 318--328. Springer, 2023.

\bibitem[Entezari et~al.(2021)Entezari, Sedghi, Saukh, and Neyshabur]{entezari2021role}
Rahim Entezari, Hanie Sedghi, Olga Saukh, and Behnam Neyshabur.
\newblock The role of permutation invariance in linear mode connectivity of neural networks.
\newblock \emph{arXiv preprint arXiv:2110.06296}, 2021.

\bibitem[Gu and Dao(2023)]{gu2023mamba}
Albert Gu and Tri Dao.
\newblock Mamba: Linear-time sequence modeling with selective state spaces.
\newblock \emph{arXiv preprint arXiv:2312.00752}, 2023.

\bibitem[Guan and Huang(2020)]{guan2020multi}
Qingji Guan and Yaping Huang.
\newblock Multi-label chest x-ray image classification via category-wise residual attention learning.
\newblock \emph{Pattern Recognition Letters}, 130:\penalty0 259--266, 2020.

\bibitem[Guendel et~al.(2019)Guendel, Grbic, Georgescu, Liu, Maier, and Comaniciu]{guendel2019learning}
Sebastian Guendel, Sasa Grbic, Bogdan Georgescu, Siqi Liu, Andreas Maier, and Dorin Comaniciu.
\newblock Learning to recognize abnormalities in chest x-rays with location-aware dense networks.
\newblock In \emph{Progress in Pattern Recognition, Image Analysis, Computer Vision, and Applications: 23rd Iberoamerican Congress, CIARP 2018, Madrid, Spain, November 19-22, 2018, Proceedings 23}, pages 757--765. Springer, 2019.

\bibitem[Jordan et~al.(2022)Jordan, Sedghi, Saukh, Entezari, and Neyshabur]{jordan2022repair}
Keller Jordan, Hanie Sedghi, Olga Saukh, Rahim Entezari, and Behnam Neyshabur.
\newblock Repair: Renormalizing permuted activations for interpolation repair.
\newblock \emph{arXiv preprint arXiv:2211.08403}, 2022.

\bibitem[Ma et~al.(2019)Ma, Wang, and Hoi]{ma2019multi}
Congbo Ma, Hu~Wang, and Steven~CH Hoi.
\newblock Multi-label thoracic disease image classification with cross-attention networks.
\newblock In \emph{Medical Image Computing and Computer Assisted Intervention--MICCAI 2019: 22nd International Conference, Shenzhen, China, October 13--17, 2019, Proceedings, Part VI 22}, pages 730--738. Springer, 2019.

\bibitem[Rajpurkar(2017)]{rajpurkar2017chexnet}
P~Rajpurkar.
\newblock Chexnet: Radiologist-level pneumonia detection on chest x-rays with deep learning.
\newblock \emph{ArXiv abs/1711}, 5225, 2017.

\bibitem[Stoica et~al.(2023)Stoica, Bolya, Bjorner, Ramesh, Hearn, and Hoffman]{stoica2023zipit}
George Stoica, Daniel Bolya, Jakob Bjorner, Pratik Ramesh, Taylor Hearn, and Judy Hoffman.
\newblock Zipit! merging models from different tasks without training.
\newblock \emph{arXiv preprint arXiv:2305.03053}, 2023.

\bibitem[Tang et~al.(2018)Tang, Wang, Harrison, Lu, Xiao, and Summers]{tang2018attention}
Yuxing Tang, Xiaosong Wang, Adam~P Harrison, Le~Lu, Jing Xiao, and Ronald~M Summers.
\newblock Attention-guided curriculum learning for weakly supervised classification and localization of thoracic diseases on chest radiographs.
\newblock In \emph{Machine Learning in Medical Imaging: 9th International Workshop, MLMI 2018, Held in Conjunction with MICCAI 2018, Granada, Spain, September 16, 2018, Proceedings 9}, pages 249--258. Springer, 2018.

\bibitem[Vaswani et~al.(2017)Vaswani, Shazeer, Parmar, Uszkoreit, Jones, Gomez, Kaiser, and Polosukhin]{vaswani2017attention}
Ashish Vaswani, Noam Shazeer, Niki Parmar, Jakob Uszkoreit, Llion Jones, Aidan~N Gomez, {\L}ukasz Kaiser, and Illia Polosukhin.
\newblock Attention is all you need.
\newblock \emph{Advances in neural information processing systems}, 30, 2017.

\bibitem[Wang et~al.(2024)Wang, Ma, Almakky, Reid, Carneiro, and Yaqub]{wang2024rethinking}
Hu~Wang, Congbo Ma, Ibrahim Almakky, Ian Reid, Gustavo Carneiro, and Mohammad Yaqub.
\newblock Rethinking weight-averaged model-merging.
\newblock \emph{arXiv preprint arXiv:2411.09263}, 2024.

\bibitem[Wang et~al.(2017)Wang, Peng, Lu, Lu, Bagheri, and Summers]{wang2017chestx}
Xiaosong Wang, Yifan Peng, Le~Lu, Zhiyong Lu, Mohammadhadi Bagheri, and Ronald~M Summers.
\newblock Chestx-ray8: Hospital-scale chest x-ray database and benchmarks on weakly-supervised classification and localization of common thorax diseases.
\newblock In \emph{Proceedings of the IEEE conference on computer vision and pattern recognition}, pages 2097--2106, 2017.

\bibitem[Wortsman et~al.(2022)Wortsman, Ilharco, Gadre, Roelofs, Gontijo-Lopes, Morcos, Namkoong, Farhadi, Carmon, Kornblith, et~al.]{wortsman2022model}
Mitchell Wortsman, Gabriel Ilharco, Samir~Ya Gadre, Rebecca Roelofs, Raphael Gontijo-Lopes, Ari~S Morcos, Hongseok Namkoong, Ali Farhadi, Yair Carmon, Simon Kornblith, et~al.
\newblock Model soups: averaging weights of multiple fine-tuned models improves accuracy without increasing inference time.
\newblock In \emph{International conference on machine learning}, pages 23965--23998. PMLR, 2022.

\bibitem[Yadav et~al.(2023)Yadav, Tam, Choshen, Raffel, and Bansal]{yadav2023ties}
Prateek Yadav, Derek Tam, Leshem Choshen, Colin~A Raffel, and Mohit Bansal.
\newblock Ties-merging: Resolving interference when merging models.
\newblock \emph{Advances in Neural Information Processing Systems}, 36:\penalty0 7093--7115, 2023.

\bibitem[Yao et~al.()Yao, Poblenz, Dagunts, Covington, Bernard, and Lyman]{yao1710learning}
Li~Yao, Eric Poblenz, Dmitry Dagunts, Ben Covington, Devon Bernard, and Kevin Lyman.
\newblock Learning to diagnose from scratch by exploiting dependencies among labels. arxiv 2017.
\newblock \emph{arXiv preprint arXiv:1710.10501}.

\end{thebibliography}
\end{document}